\documentclass[conference]{IEEEtran}
\IEEEoverridecommandlockouts
\usepackage{cite}
\usepackage{amsmath,amssymb,amsfonts}
\usepackage{algorithmic}
\usepackage{graphicx}
\usepackage{textcomp}
\usepackage{xcolor}
\def\BibTeX{{\rm B\kern-.05em{\sc i\kern-.025em b}\kern-.08em
    T\kern-.1667em\lower.7ex\hbox{E}\kern-.125emX}}

\usepackage[utf8]{inputenc}
\usepackage{mathtools}
\usepackage{amsthm, color}
\usepackage{bm, makecell}
\usepackage{tikz, subcaption, cite}
\usepackage{multirow}
\usepackage{array,booktabs,arydshln,xcolor}
\graphicspath{{fig/}}

\begin{document}

\title{Conditional Variable Selection for Intelligent Test
\thanks{This research was supported by Advantest as part of the Graduate School ``Intelligent Methods for Test and Reliability'' (GS-IMTR) at the University of Stuttgart. We also appreciate Zahra Paria Najafi Haghi for providing the simulation datasets for combiantional circuits.}
}


\author{\IEEEauthorblockN{Yiwen Liao\IEEEauthorrefmark{1}, Tianjie Ge\IEEEauthorrefmark{1}, Rapha\"el Latty\IEEEauthorrefmark{2}, and Bin Yang\IEEEauthorrefmark{1}}
	\IEEEauthorblockA{\IEEEauthorrefmark{1}Institute of Signal Processing and System Theory, University of Stuttgart, Germany\\
		Email: \{yiwen.liao,bin.yang\}@iss.uni-stuttgart.de, tianjie.ge.ariel@gmail.com}
	\IEEEauthorblockA{\IEEEauthorrefmark{2}Applied Research and Venture Team, Advantest Europe GmbH, Germany\\
		Email: raphael.latty@advantest.com}
}

\maketitle

\begin{abstract}
	Intelligent test requires efficient and effective analysis of high-dimensional data in a large scale. Traditionally, the analysis is often conducted by human experts, but it is not scalable in the era of big data. To tackle this challenge, variable selection has been recently introduced to intelligent test. However, in practice, we encounter scenarios where certain variables (e.g. some specific processing conditions for a device under test) must be maintained after variable selection. We call this conditional variable selection, which has not been well investigated for embedded or deep-learning-based variable selection methods. In this paper, we discuss a novel conditional variable selection framework that can select the most important candidate variables given a set of preselected variables.
\end{abstract}


\section{Introduction}
\label{sec:introduction}

%

Test and reliability play a critical role throughout the entire semiconductor industry from chip design to mass production. With the fast development of the semiconductor industry, however, large-scale data with high dimensionality have become inevitable. This is accordingly one of the major bottlenecks for efficient and intelligent test. For example, data of high dimensionality are difficult to visualize in a 2D- or 3D-space and thus demanding for domain experts to explore the data. Also, due to the curse of dimensionality, modeling the relations between candidate and target variables by statistical approaches is problematic. Thereby, in order to tackle these problems, variable selection~\cite{guyon2003introduction} has been introduced to data analysis for test and reliability in our recent research~\cite{liao2022tuz}. In particular, we designed a novel variable selection approach called FM-module~\cite{9533531} based on Deep Learning (DL) and used it to identify a few most critical and representative candidate variables with respect to the given target variables. In this way, the data dimensionality is significantly reduced with high efficiency and effectiveness.

Despite the advancement and success of applying our approach to reduce data dimensionality for intelligent test, the proposed method simultaneously considers all candidate variables without taking any additional information during training into account. On the contrary, in practice, certain candidate variables must be maintained after selection due to predefined requirements or expert knowledge. For example, delays under the nominal voltage are considered as a critical characteristic for a circuit, so the tests under the nominal voltage must be maintained after selection. Unfortunately, the existing method has not considered this situation yet. More precisely, to the best of our knowledge, almost no previous (embedded and DL-based) feature selection approaches have meditated a solution from this aspect.

In order to address the challenges described above, this paper proposes a generic framework, inherited from the FM-module, for conditional variable selection based on DL. Specifically, we separately encode the preselected and candidate variables into suitable representations, and then feed them to the same neural network for a given learning task. In this way, the FM-module, encoding blocks and the neural network are jointly trained. Finally, after training, we obtain a variable importance vector as the original FM-module does, each element of which denotes the importance of the corresponding candidate variable given the preselected variables.
\section{Methodology}
\label{sec:method}
\begin{figure}
	\centering
	\begin{tikzpicture}[scale=0.75]
		
		\node (xc) at(0, 0) {$X_c$};
		\node (xp) at(0, 2) {$X_p$};
		\node (xm) at(3, 0) {$X_c\odot\bm{m}$};
		\node (ocross) at(7, 1) {$\boxtimes$};
		
		\node[thick, rectangle, rounded corners=1mm, draw=gray] (fp) at (5.5, 2) {$f_p(\cdot)$};
		
		\node[thick, rectangle, rounded corners=1mm, draw=gray] (fc) at (5.5, 0) {$f_c(\cdot)$};
		
		\node[thick, rectangle, rounded corners=1mm, draw=gray] (fm) at (1.25, -0.75) {FM};
		
		\draw[thick, rounded corners=1mm, color=gray] (8, 0)--(10, 0.5)--(10, 1.5)--(8, 2.)--cycle;
		\node (g) at(9, 1) {$g(\cdot)$};
		\node (y) at(11, 1) {$\hat{Y}$};
		
		\draw[color=gray, ->] (xc)--(xm);
		\draw[color=gray, ->] (xm)--(fc);
		\draw[color=gray, ->] (xp)--(fp);
		\draw[color=gray, rounded corners=1mm, ->] (xc)--(0, -0.75)--(fm);
		\draw[color=gray, rounded corners=1mm, ->] (fm)--(3.6, -0.75)--(3.6, -0.3);
		
		\draw[color=gray, rounded corners=1mm, ->] (fp)--(7, 2)--(ocross);
		\draw[color=gray, rounded corners=1mm, ->] (fc)--(7, 0)--(ocross);
		\draw[color=gray, ->] (ocross)--(8, 1);
		\draw[color=gray, ->] (10, 1)--(y);

	\end{tikzpicture}
	\caption{The proposed generic framework for conditional variable selection.}
	\label{fig:conditional_fs}
\end{figure}
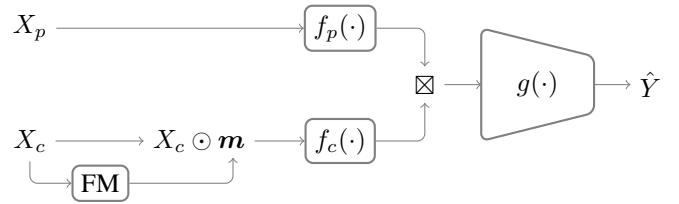

To enable conditional variable selection in neural networks, we propose to fuse the information from the preselected and candidate variables in the shallow layers of a neural network targeting a given learning task. As a result, the neural network acts as a guide to the variable selection module in a way that candidate variables, which can minimize the training losses given the preselected variables, should be assigned with greater importance scores.

In the subsequent sections, we use the following notations. The input data are denoted as a matrix $X=[\bm{x}_1, \bm{x}_2, \dots, \bm{x}_N]^T\in\mathbb{R}^{N\times D}$, where $N$ is the number of data points and $D$ is the number of variables. According to user specifications, $X$ is composed of two parts as $X=[X_p, X_n]$. That is, the \underline{p}reselected $D_p$ variables $X_p\in\mathbb{R}^{N\times D_p}$ and the $D_c$ \underline{c}andidate variables $X_c\in\mathbb{R}^{N\times D_c}$ with $D = D_c + D_p$. For a supervised variable selection, the labels (target variables) are denoted as $Y=[\bm{y}_1, \bm{y}_2, \dots, \bm{y}_N]^T\in\mathbb{R}^{N\times D_y}$, where $D_y$ is defined by the given learning task. For example, $D_y=1$ for univariate regression, and $D_y=2$ for binary classification with one-hot coding representations for class labels.

Based on the notations above, the overall framework is illustrated in Fig.~\ref{fig:conditional_fs}. The candidate variables $X_c$ are fed into the FM-module~\cite{9533531}, which generates a feature mask vector $\bm{m}\in\mathbb{R}^{D_c}$. As stated in~\cite{9533531}, each element in $\bm{m}$ is bounded by the range between 0 and 1 and denotes the importance of the corresponding candidate variable in $X_c$. Accordingly, we obtain the weighted candidate variables $X_c\odot\bm{m}$. Afterwards, a transformation $f_c(\cdot)$ is applied to $X_c\odot\bm{m}$. In parallel, another independent transformation $f_p(\cdot)$ is applied to the preselected variables $X_p$. Next, denoted by $\boxtimes$ in Fig.~\ref{fig:conditional_fs}, we fuse the outputs from both transformations by concatenating them together. The fused representations are fed to a neural network $g(\cdot)$ to obtain predictions $\hat{Y}$. The learning objective is dependent on the concrete task, e.g. a Mean Squared Error (MSE) loss can be used for univariate regression and a binary cross entropy loss can be used for binary classification. Once the full model (i.e. FM, $f_p$, $f_c$ and $g$) is trained, we calculate $\bm{m}$ based on the entire training data to obtain the final importance scores for the candidate variables only.

\section{Experiments}
\label{sec:exp}
%
%
\begin{figure}
	\centering
	\includegraphics[width=.85\linewidth]{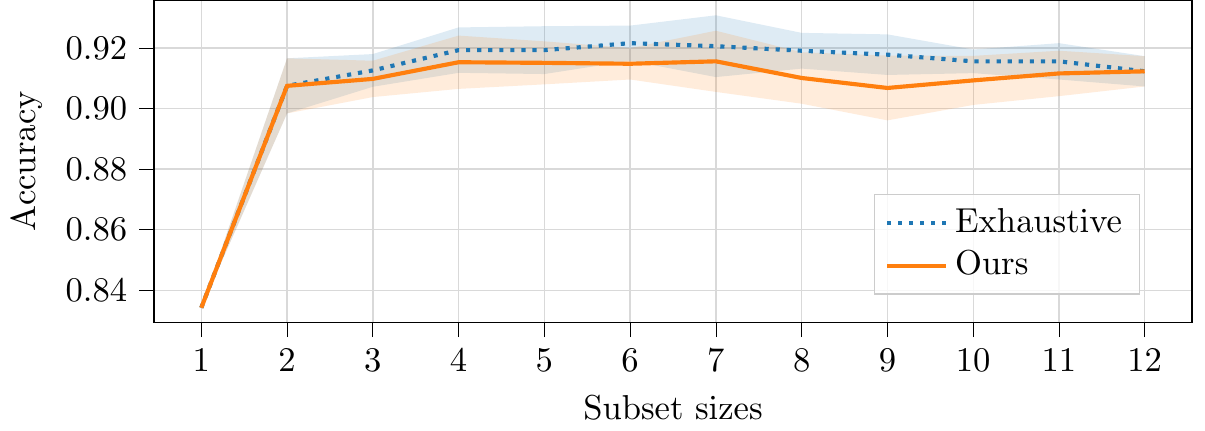}
	\caption{Accuracy over different subset sizes for our method and an exhaustive search.}
	\label{fig:ap1_cfm}
\end{figure}

\subsection{Case I: Voltage Selection for Open Circuit Detection}
The delays under varying voltages are used to identify open resistive defects from slow behavior due to process variations in combinational circuits~\cite{9651857}. This experiment investigates delays under which variables (varying voltages) are critical for defect identification given the nominal voltage of 0.9V. In total, we had respectively 1000 defective and 1000 non-defective samples with $D_c=11$ and $D_p=1$. Fig.~\ref{fig:ap1_cfm} shows the resulting accuracy over different subset sizes $K$ from 1 to 12. $K=1$ means that we selected 0.9V as the only variable, while the other sizes denote that we select 0.9V and $K-1$ different voltages. Overall, our method shows similar performance as an exhaustive search, while the exhaustive search is not feasible in practice due to its exponential computational complexity.

\subsection{Case II: Variable Selection for Post-Silicon Tuning}
\begin{figure}
	\centering
	\includegraphics[width=\linewidth]{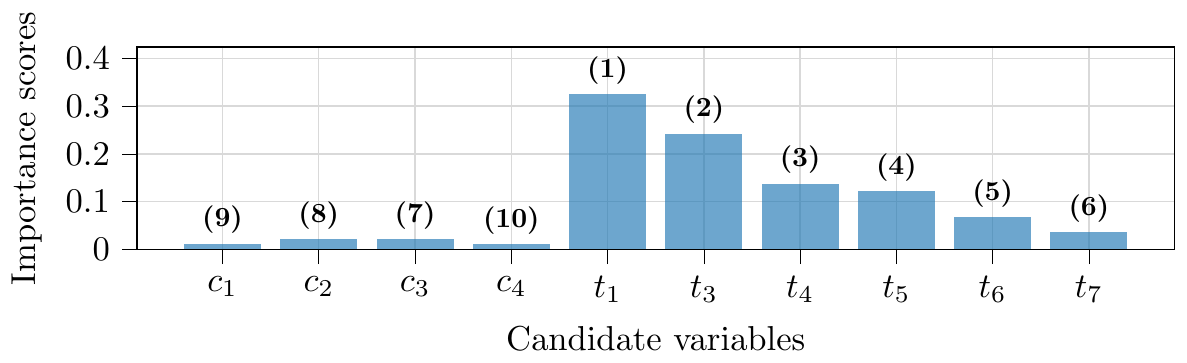}
	\caption{The learned importance scores of our method for a real-world dataset. Larger scores indicate higher importance of the corresponding candidate variables.}
	\label{fig:tuning}
\end{figure}

In this experiment, we focus on a real-world dataset from Advantest, which consists of 100,000 test cases obtained from a single Device Under Test (DUT). Each test case (sample) has 11 variables, namely $c_1$ to $c_4$ and $t_1$ to $t_7$. The goal was to find out the four most important candidate variables under the condition that $t_2$ was already preselected, i.e. $D_c=10$ and $D_p=1$. For the exhaustive search, we need to evaluate $\binom{10}{4}=210$ different candidate variable combinations and trained a 3-layer multilayer perceptron by minimizing the MSE loss between the predictions and the univariate target variable, i.e. the Figure-of-Merit (FoM). In total, it took about 86 minutes for the exhaustive search and the optimal selection result was the combination of $t_1$ to $t_5$. On the contrary, our method was trained only once by feeding all 10 candidate variables and preselected variables to the model. The learned variable importance vector is shown in Figure.~\ref{fig:tuning}. We can see that $t_1$, $t_3$, $t_4$ and $t_5$ obtained the greatest importance scores and were considered as the best candidate variables given the preselected variable $t_2$, which matched the aforementioned exhaustive search results. It should be emphasized that it took about only 25 seconds (i.e. 0.4 minutes) to perform the full training for our method, corresponding to about 0.48\% time consumption of an exhaustive search with the same learning network architecture. The experimental results show that our method is promising to efficiently solve conditional variable selection tasks in test and reliability.

\section{Conclusion}
\label{sec:conclusion}
This paper proposes a generic framework for conditional variable selection based on deep learning by encoding the preselected variables and fuse the resulting representations and those of the candidate variables. Preliminary experiments on both simulation and real-world datasets have shown promising results. Future research may include exploration of different encoding approaches and deeper neural network architectures.

\bibliographystyle{IEEEtran}
\bibliography{Refs}

\begin{thebibliography}{1}
\providecommand{\url}[1]{#1}
\csname url@samestyle\endcsname
\providecommand{\newblock}{\relax}
\providecommand{\bibinfo}[2]{#2}
\providecommand{\BIBentrySTDinterwordspacing}{\spaceskip=0pt\relax}
\providecommand{\BIBentryALTinterwordstretchfactor}{4}
\providecommand{\BIBentryALTinterwordspacing}{\spaceskip=\fontdimen2\font plus
\BIBentryALTinterwordstretchfactor\fontdimen3\font minus
  \fontdimen4\font\relax}
\providecommand{\BIBforeignlanguage}[2]{{%
\expandafter\ifx\csname l@#1\endcsname\relax
\typeout{** WARNING: IEEEtran.bst: No hyphenation pattern has been}%
\typeout{** loaded for the language `#1'. Using the pattern for}%
\typeout{** the default language instead.}%
\else
\language=\csname l@#1\endcsname
\fi
#2}}
\providecommand{\BIBdecl}{\relax}
\BIBdecl

\bibitem{guyon2003introduction}
I.~Guyon and A.~Elisseeff, ``An introduction to variable and feature
  selection,'' \emph{Journal of machine learning research}, vol.~3, no. Mar,
  pp. 1157--1182, 2003.

\bibitem{liao2022tuz}
Y.~Liao, J.~Rivoir, R.~Latty, and B.~Yang, ``A deep-learning-aided pipeline for
  efficient post-silicon tuning,'' in \emph{34. Workshop Testmethoden und
  Zuverl\"assigkeit von Schaltungen und Systemen}, 2022.

\bibitem{9533531}
Y.~Liao, R.~Latty, and B.~Yang, ``Feature selection using batch-wise
  attenuation and feature mask normalization,'' in \emph{2021 International
  Joint Conference on Neural Networks (IJCNN)}, 2021, pp. 1--9.

\bibitem{9651857}
Z.~P. Najafi-Haghi and H.-J. Wunderlich, ``Resistive open defect classification
  of embedded cells under variations,'' in \emph{2021 IEEE 22nd Latin American
  Test Symposium (LATS)}, 2021, pp. 1--6.

\end{thebibliography}

\end{document}